\documentclass[10pt,twocolumn]{article}

\usepackage[]{emnlp2021}
\usepackage{times}
\usepackage{latexsym}
\usepackage{arydshln}
\usepackage{fixltx2e}
\usepackage{graphicx}
\usepackage{pgfplots}
\usepackage{caption}
\usepackage[normalem]{ulem}

\usepackage{microtype}

\title{Noisy Text Data: Achilles' Heel of popular transformer based NLP models}
\author{Kartikay Bagla\textsuperscript{1}, Ankit Kumar\textsuperscript{2}, Shivam Gupta\textsuperscript{3}, Anuj Gupta\textsuperscript{2}  \\ \textsuperscript{1}Delhi Technological University \\ \textsuperscript{2}Vahan Inc. \\ \textsuperscript{3} Ninja Salary }

\begin{document}

\maketitle

\begin{abstract}
In the last few years, the ML community has created a number of new NLP models based on transformer architecture. These models have shown great performance for various NLP tasks on benchmark datasets, often surpassing SOTA results. Buoyed with this success, one often finds industry practitioners actively experimenting with fine-tuning these models to build NLP applications for industry use cases. However, for most datasets that are used by practitioners to build industrial NLP applications, it is hard to guarantee the presence of any noise in the data. While most transformer based NLP models have performed exceedingly well in transferring the learnings from one dataset to another, it remains unclear how these models perform when fine-tuned on noisy text.

We address the open question by  \newcite{kumar-etal-2020-noisy} to explore the sensitivity of popular transformer based NLP models to noise in the text data. We continue working with the noise as defined by them - spelling mistakes \& typos (which are the most commonly occurring noise). We show (via experimental results) that these models perform badly on most common NLP tasks namely text classification, textual similarity, NER, question answering, text summarization on benchmark datasets. We further show that as the noise in data increases, the performance degrades. Our findings suggest that one must be vary of the presence of noise in their datasets while fine-tuning popular transformer based NLP models.





\end{abstract}

\section{Introduction}
It is a well known fact that pre-trained contextualized language models such as BERT (Bidirectional Encoder Representations from Transformers) \newcite{devlin2018bert}, BART (Bidirectional and Auto-Regressive Transformer) \newcite{lewis2019bart}, RoBERTa (robustly optimized BERT pretraining approach) \newcite{liu2019roberta}, ALBERT (A Lite BERT) \newcite{lan2019albert}, XLNet (Generalized autoregressive pretraining for language understanding) \newcite{yang2019xlnet}, T5 (Text-to-Text Transfer Transformer) \newcite{raffel2019exploring} have shown remarkable gains in performance for various Natural Language Processing (NLP) tasks. This includes most common downstream tasks such as Text Classification, Textual Similarity, Summarization, Name-Entity Recognition, Question Answering, Machine Translation etc.

Given this fantastic progress, machine learning teams in the industry are actively experimenting with fine-tuning these models on their data to solve industry use cases. These include applications such as chatbots, sentiment analysis systems, intelligent ticketing systems, entity recognition systems, machine translation systems etc. To building these applications practitioners often assemble the required dataset by collecting text data from data sources \& applications such as chats, emails, discussions from user forums, social media conversations, output of machine translation systems, automatically transcribing text from speech data, automatically recognized text from printed or handwritten material, etc. The text data from such applications is often noisy (The amount of noise may differ depending on the source). For example, the data coming from discussions on user forums \& social media conversations, the noise in the text data can be significantly high.

\newcite{kumar-etal-2020-noisy} shows that the performance of pre-trained BERT degrades significantly when fine-tuned on noisy text data. We extend their work to show this holds true for most popular transformer based NLP models, namely BERT, BART, RoBERTa, ALBERT, XLNet and T5. To be precise Kumar et. al \cite{} benchmarked the performance of BERT when fine tuned on noisy text data for 2 tasks - text classification and textual similarity using 3 datasets - IMDB, SST-2 and STS-B. The main contributions of this paper are three fold: 
\begin{itemize}
\item We extend the work of \newcite{kumar-etal-2020-noisy} to benchmark the performance of BERT when fine tuned on noisy text data for other fundamental NLP tasks - Question Answering, NER and Summerization using SQUAD, CoNLL and Billsum datasets respectively. 
\item We benchmark the performance of other popular transformer based NLP models - BART, RoBERTa, ALBERT, XLNet and T5 on fundamental NLP tasks - text classification, textual similarity, Question Answering, NER and Summerization on noisy text data\footnote{Certain models are not applicable for certain tasks. We skip those tasks}. 
\item We show that all the above mentioned models perform badly when fine-tuned on noisy text data. Further, as the noise increases, performance becomes worse.
\end{itemize}

This work is motivated from a business use case where we built a conversational system over WhatsApp to screen job seekers for blue collar jobs. The candidates often are not even college graduates. This paired with the fat finger problem\footnote{https://en.wikipedia.org/wiki/Fat-finger\_error} over a smartphone keypad often results in many typos and spelling mistakes in the responses job seekers send to our conversational system. Though this work is inspired from our business use case, our findings are applicable to other use cases that deal with noisy text data. 




\section{Previous Work}
Modern communication mediums such as SMS, chats, twitter, messaging apps encourage brevity and informalism, leading to non-canonical text. This presents significant challenges to the known NLP techniques. Research community has done a lot of work on validating various transformer based  language models in the presence of noisy text. The NLP community has done a lot of work on understanding the effects of noise on the performance of NLP models. \newcite{taghva2000evaluating} evaluate the effect of OCR errors on text categorization. \newcite{wu2016google} introduced ISSAC, a system to clean dirty text from online sources. \newcite{belinkov2017synthetic} show that character basedneural machine translation (NMT) models are also prone to synthetic and natural noise even though these model dobetter job to handle out-of-vocabulary issues and learn better morphological representation.

~\newcite{aspillaga2020stress} evaluated RoBERTa, XLNet, and BERT in Natural Language Inference (NLI) and Question Answering (QA) tasks. They used BiDAF~\cite{seo2016bidirectional} and Match-LSTM~\cite{wang2016machine} as baselines to compare stress tests against Transformer-based models. They did two type of tests - distraction test and noise test(spelling errors). They show that RoBERTa, XLNet and BERT are more robust in stress tests than recurrent neural network models.~\newcite{ravichander2021noiseqa} describes a real world scenario where question answering system can be affected by different types of noise such as keyboard errors and ASR errors. They evaluate SOTA methods on natural and synthetic noisy data and demonstrate that the performance of QA systems is impacted by the real world noise. They further analyze synthetic noise and its impact on the downstream question answering system and presented an initial exploration of mitigation strategies for real world noise.
\newcite{alshemali2020improving} test DNN models on different NLP tasks like Classification, Machine Translation, Question Answering, Textual Entailment, Tagging etc. They illustrate the vulnerability of DNNs to adversarial examples — inputs modified by introducing small perturbations to deliberately fool the target model into giving incorrect results. \newcite{agarwal2007much} study the effect of different kinds of noise on automatic Text Classification. They present detailed experimental results with simulated noise on the Reuters21578 and 20-newsgroups benchmark datasets; also with results on real-life noisy datasets from various CRM domains. They use spelling errors to generate synthetic dataset and show the effect of noise on final accuracy of text classification task.~\newcite{Subramaniam2009} present a survey on the types of text noise and techniques to handle it.  \newcite{belinkov2017synthetic} show that character based neural machine translation (NMT) models are prone to synthetic and natural noise even though these model do better job to handle out-of-vocabulary issues and learn better morphological representation.~\newcite{ribeiro2018semantically} developed a technique, called semantically equivalent adversarial rules (SEARs) to debug NLP models. SEAR generates adversial examples to penetrate NLP models. Authors experimented this techniques for three domains: machine comprehension, visual question answering, and sentiment analysis.\\
\newcite{Sun2020AdvBERTBI} explored robustness of BERT in dealing with noisy data. They experimented with sentiment analysis and question answering tasks. They inject keyboard typos in data in two ways - a) adding typos randomly b) adding typos only in the most informative words. They show that typos in informative words cause severer degradation.~\newcite{pal2020transfer} presented novel attack techniques that utilized the unintended features learnt in the \textit{teacher} (public) model to generate adversarial examples for \textit{student} (downstream) models. They show that using length-based and sentence-based misclassification attacks for the Fake News Detection task trained using a context-aware BERT model, one gets misclassification accuracy of 78\% and 39\% respectively for the adversarial examples.~\newcite{jin2019bert} introduced TEXTFOOLER, a system which generates adversarial text and applied it to text classification and textual entailment to successfully attack the pre-trained BERT among other models.

\section{Experiments}
We evaluate the 6 transformer based NLP models - BERT\footnote{BERT\textsubscript{Base} uncased model}, BART, RoBERTa, ALBERT, XLNet and T5. For this evaluation we use the following tasks [and corresponding datasets]: 
\begin{enumerate}
\item Sentiment Analysis [IMDB movie reviews \cite{maas2011learning} and Stanford Sentiment Treebank (SST-2)~\cite{socher2013recursive}]
\item Textual similarity [Semantic Textual Similarity (STS-B) \cite{cer2017semeval}]
\item Question Answering [SQuAD2.0 \cite{squad2.0}]
\item Named Entity Recognition [CoNLL \cite{sang2003introduction} ]
\item Text summerization [Billsum \cite{billsumdataset}]

\end{enumerate}

\noindent On each of these 6 datasets, we report the performance of each of the 6 models\footnote{Certain models are not applicable for certain tasks. We skip those.}.  both - with and without noise. 

\subsection{Noise}

We directly borrow the notion of noise as defined by \cite{kumar-etal-2020-noisy}.
They focus on the noise introduced by spelling mistakes and typos. All the benchmark datasets used consist of examples X \(\rightarrow\) Y where X is the text input and Y is the corresponding label. They call the original dataset as D\textsubscript{0}. From D\textsubscript{0} they create new datasets  D\textsubscript{5},  D\textsubscript{10},  D\textsubscript{15},  D\textsubscript{20} and D\textsubscript{25}. D\textsubscript{k} is a variant of D\textsubscript{0} with k\% noise in each datapoint in D\textsubscript{0}. 
 
To create D\textsubscript{k}, they take i\textsuperscript{th} data point x\textsubscript{i} \(\in\) D\textsubscript{k}, and introduce noise in it. They represent the modified datapoint by x\textsubscript{i,k}\textsuperscript{noise} . Then, D\textsubscript{k} is simply the collection (x\textsubscript{i,k} \textsuperscript{noise}, y\textsubscript{i}), \(\forall\)i. To create x\textsubscript{i,k}\textsuperscript{noise} from x\textsubscript{i}, they randomly choose k\% characters from the text of x\textsubscript{i} and replace them with nearby characters in a qwerty keyboard. For example, if character \emph{d} is chosen, then it is replaced by a character randomly chosen from \emph{e}, \emph{s}, \emph{x}, \emph{c}, \emph{f}, or \emph{r}. This is because in a qwerty keyboard, these keys surround the key \emph{d}. They  inject noise in the complete dataset. Once  split D\textsubscript{i} into \textit{train} and \textit{test} chunks.

\subsection{Text Classification}
For text classification we use IMDB movie reviews \cite{maas2011learning} and Stanford Sentiment Treebank (SST-2 ) \cite{socher2013recursive} datasets in binary prediction settings. IMDB datasets consist of 25000 training and 25000 test sentences. We represent the original IMDB dataset (one with no noise) as IMDB\textsubscript{0}. Using the process of introducing noise (as described in section 3.1), we create 5 variants of IMDB\textsubscript{0} namely IMDB\textsubscript{5}, \ldots, IMDB\textsubscript{25} with varying degrees of noise. 

SST-2 dataset consists of 67349 training and 872 test sentences. Here too we we add noise as described in Section 3.1 to create 5 variants of SST-2\textsubscript{0} -  SST-2\textsubscript{5}, \ldots,  SST-2\textsubscript{25}.
To measure the performance for text classification, we use F1 score. 

\subsection{Textual Similarity}

For the textual similarity task, we use the Semantic Textual Similarity (STS-B) \cite{cer2017semeval} dataset. The dataset consists of 5749 training and 1500 test data points. Each data point consists of 2 sentences and a score between 0-5 representing the similarity between the two sentences. We represent the original data set by STS-B\textsubscript{0} and create 5 noisy variants.
Here, we use Pearson-Spearman correlation to measure model's performance.

\subsection{Question Answering}
For the question answering task, we use the Stanford Question Answering Dataset version 2.0 (SQUAD2.0)\cite{squad2.0}. This dataset has 129,941 training and 5915 test paragraph and question pairs. The evaluation metric we used is F1. F1 score takes each gold answer as bags of words and doesn’t require choosing the exact same
span as a human’s, which is seen as more reliable. 

\subsection{Named Entity Recognition}
We used the CoNLL-2003\cite{sang2003introduction} dataset for Named Entity Recognition. It consists of 22,137 sentences
totally and is split into 14,987,and 3,684 sentences for the training and test sets, respectively. It is tagged with four linguistic entity types (PER, LOC, ORG, MISC). We used average F1 score as evaluation metric. 

\subsection{Text Summarization}
For the text summarization task, we used the BillSum\cite{billsumdataset} dataset. It contains 23,000 US Congressional bills and human-written reference summaries from the 103rd-115th (1993-2018) sessions of Congress. We use ROUGE1-F1(\newcite{rouge2004package}) metric for the summarization task.

\begin{table}[]
\small
\begin{tabular}{c|c|c|c|c}
\hline
\multicolumn{5}{c}{\textbf{IMDB}} \\    
\hline
\textbf{\% error} & \textbf{BERT} & \textbf{RoBERTa} & \textbf{ALBERT} & \textbf{XLNet}              \\
\hdashline
0 & 0.938 & 0.947 & 0.935 & 0.948\\
5 & 0.914 & 0.929 & 0.908 & 0.922\\
10 & 0.882 & 0.905 & 0.865 & 0.894\\
15 & 0.843 & 0.872 & 0.782 & 0.859\\
20 & 0.81 & 0.835 & 0.781 & 0.83\\
25 & 0.758 & 0.805 & 0.77 & 0.775
\end{tabular}
\vspace{10pt}
\caption{F1 scores on IMDB. }
\label{table:IMDB}
\end{table}

\begin{table}[]
\small
\begin{tabular}{c|c|c|c|c}
\hline
\multicolumn{5}{c}{\textbf{SST-2}} \\    
\hline
\textbf{\% error} & \textbf{BERT} & \textbf{RoBERTa} & \textbf{ALBERT} & \textbf{XLNet}              \\
\hdashline
0 & 0.918 & 0.933 & 0.907 & 0.913\\
5 & 0.88 & 0.893 & 0.856 & 0.885\\
10 & 0.847 & 0.865 & 0.802 & 0.833\\
15 & 0.804 & 0.827 & 0.763 & 0.807\\
20 & 0.79 & 0.781 & 0.76 & 0.754\\
25 & 0.746 & 0.733 & 0.689 & 0.721
\end{tabular}
\vspace{10pt}
\caption{F1 scores on SST-2.}
\label{table:SST}
\end{table}

\begin{table}[]
\small
\begin{tabular}{c|c|c|c|c}
\hline
\multicolumn{5}{c}{\textbf{STS-B}} \\
\hline
\textbf{\% error} & \textbf{BERT} & \textbf{RoBERTa} & \textbf{ALBERT} & \textbf{XLNet}              \\
\hdashline
0 & 0.896 & 0.906 & 0.89 & 0.883\\
5 & 0.7794 & 0.848 & 0.766 & 0.746\\
10 & 0.624 & 0.761 & 0.59 & 0.571\\
15 & 0.49 & 0.668 & 0.388 & 0.449\\
20 & 0.398 & 0.513 & 0.232 & 0.336\\
25 & 0.355 & 0.497 & 0.252 & 0.288
\end{tabular}
\vspace{10pt}
\caption{Pearson-Spearman correlations on STS-B.}
\label{table:STS-B}
\end{table}

\begin{table}[]
\small
\begin{tabular}{c|c|c|c}
\hline
\multicolumn{4}{c}{\textbf{SQUAD2}} \\    
\hline
\textbf{\% error} & \textbf{BERT} & \textbf{RoBERTa} & \textbf{ALBERT}          
\\
\hdashline
0 & 71.8 & 82.36 & 78.01\\
5 & 64.98 & 77.04 & 70.38\\
10 & 55.16 & 68.6 & 60.33\\
15 & 48.18 & 61.74 & 53.02\\
20 & 41.7 & 54.08 & 46.03\\
25 & 33 & 48.16 & 39.17
\end{tabular}
\vspace{10pt}
\caption{F1 scores on SQUAD2. }
\label{table:SQUAD2}
\end{table}

\begin{table}[]
\small
\begin{tabular}{c|c|c|c|c}
\hline
\multicolumn{5}{c}{\textbf{CoNLL}} \\    
\hline
\textbf{\% error} & \textbf{BERT} & \textbf{RoBERTa} & \textbf{ALBERT} & \textbf{XLNet}              \\
\hdashline
0 & 94.49 & 96.05 & 93.35 & 95.85\\
5 & 90 & 93.62 & 87.17 & 92.23\\
10 & 84.96 & 91.18 & 83.41 & 90.02\\
15 & 81.19 & 88.47 & 78.8 & 87.23\\
20 & 78.42 & 85.88 & 74.48 & 84.28\\
25 & 75.25 & 83.6 & 70.75 & 81.88
\end{tabular}
\vspace{10pt}
\caption{F1 scores on CoNLL. }
\label{table:CoNLL}
\end{table}

\begin{table}[]
\small
\begin{tabular}{c|c|c|c|c}
\hline
\multicolumn{5}{c}{\textbf{BillSum}} \\    
\hline
\textbf{\% error} & \textbf{t5-small} & \textbf{bart-base} & \textbf{distil-bart} & \textbf{t5-base}              \\
\hdashline
0 & 57.01 & 57.76 & 57.99 & 60.82\\
5 & 53.23 & 56.01 & 55.84 & 57.66\\
10 & 49.76 & 54.85 & 54.46 & 55.33\\
15 & 47.49 & 53.07 & 53.41 & 53.39\\
20 & 45.54 & 52.52 & 52.39 & 51.76\\
25 & 43.76 & 51.41 & 51.41 & 49.98
\end{tabular}
\vspace{10pt}
\caption{ ROUGE1-F1 scores on BillSum.}
\label{table:billsum}
\end{table}

\subsection{Results}
The results of various experiments are shown in Tables ~\ref{table:IMDB}, ~\ref{table:SST}, ~\ref{table:STS-B}, ~\ref{table:SQUAD2}, ~\ref{table:CoNLL} and ~\ref{table:billsum} respectively. Note that 0\% error case represents a no noise scenario. Interestingly for most scenarios across tasks - RoBERTa consistently gives better performance as compared to other three models for same amount of noise. At the same time BERT and ALBERT show bad performance.  

\section{Conclusion and Future Work}
In this work, we studied the effect of synthetic noise (spelling mistakes) in text data on the performance of popular transformer based language models. Our experiments show that as the noise in the data increases, model performance drops significantly. Our work shows that one must be congnizant of the presence of any noise in their text data if they are fine tuning NLP models on noisy text data. Further, if there is noise is text data, then either 1) one has to preprocess data until all noise is removed. This can become a full fledged project in its own. 2) make changes to the architecture of these models to make them robust to noise. We leave this as a future work.

It will also be interesting to see how these models perform in the presence of other types of noise. It also remains to be seen if the results will hold when the noise is restricted to only frequent misspellings. Also it remains to be seen why RoBERTa shows more stability to noise unlike BERT and ALBERT.    


\bibliographystyle{acl_natbib}
\bibliography{NLP_SOTA}

\begin{thebibliography}{28}
\expandafter\ifx\csname natexlab\endcsname\relax\def\natexlab#1{#1}\fi

\bibitem[{Agarwal et~al.(2007)Agarwal, Godbole, Punjani, and
  Roy}]{agarwal2007much}
Sumeet Agarwal, Shantanu Godbole, Diwakar Punjani, and Shourya Roy. 2007.
\newblock How much noise is too much: A study in automatic text classification.
\newblock In \emph{Seventh IEEE International Conference on Data Mining (ICDM
  2007)}, pages 3--12. IEEE.

\bibitem[{Alshemali and Kalita(2020)}]{alshemali2020improving}
Basemah Alshemali and Jugal Kalita. 2020.
\newblock Improving the reliability of deep neural networks in nlp: A review.
\newblock \emph{Knowledge-Based Systems}, 191:105210.

\bibitem[{Aspillaga et~al.(2020)Aspillaga, Carvallo, and
  Araujo}]{aspillaga2020stress}
Carlos Aspillaga, Andr{\'e}s Carvallo, and Vladimir Araujo. 2020.
\newblock Stress test evaluation of transformer-based models in natural
  language understanding tasks.
\newblock \emph{arXiv preprint arXiv:2002.06261}.

\bibitem[{Belinkov and Bisk(2017)}]{belinkov2017synthetic}
Yonatan Belinkov and Yonatan Bisk. 2017.
\newblock Synthetic and natural noise both break neural machine translation.
\newblock \emph{arXiv preprint arXiv:1711.02173}.

\bibitem[{Cer et~al.(2017)Cer, Diab, Agirre, Lopez-Gazpio, and
  Specia}]{cer2017semeval}
Daniel Cer, Mona Diab, Eneko Agirre, Inigo Lopez-Gazpio, and Lucia Specia.
  2017.
\newblock Semeval-2017 task 1: Semantic textual similarity-multilingual and
  cross-lingual focused evaluation.
\newblock \emph{arXiv preprint arXiv:1708.00055}.

\bibitem[{Devlin et~al.(2018)Devlin, Chang, Lee, and
  Toutanova}]{devlin2018bert}
Jacob Devlin, Ming-Wei Chang, Kenton Lee, and Kristina Toutanova. 2018.
\newblock Bert: Pre-training of deep bidirectional transformers for language
  understanding.
\newblock \emph{arXiv preprint arXiv:1810.04805}.

\bibitem[{Jin et~al.(2019)Jin, Jin, Zhou, and Szolovits}]{jin2019bert}
Di~Jin, Zhijing Jin, Joey~Tianyi Zhou, and Peter Szolovits. 2019.
\newblock Is bert really robust.
\newblock \emph{A Strong Baseline for Natural Language Attack on Text
  Classification and Entailment}.

\bibitem[{Kornilova and Eidelman(2019)}]{billsumdataset}
Anastassia Kornilova and Vlad Eidelman. 2019.
\newblock \href {http://arxiv.org/abs/1910.00523} {Billsum: {A} corpus for
  automatic summarization of {US} legislation}.
\newblock \emph{CoRR}, abs/1910.00523.

\bibitem[{Kumar et~al.(2020)Kumar, Makhija, and Gupta}]{kumar-etal-2020-noisy}
Ankit Kumar, Piyush Makhija, and Anuj Gupta. 2020.
\newblock \href {https://doi.org/10.18653/v1/2020.wnut-1.3} {Noisy text data:
  Achilles{'} heel of {BERT}}.
\newblock In \emph{Proceedings of the Sixth Workshop on Noisy User-generated
  Text (W-NUT 2020)}, pages 16--21, Online. Association for Computational
  Linguistics.

\bibitem[{Lan et~al.(2019)Lan, Chen, Goodman, Gimpel, Sharma, and
  Soricut}]{lan2019albert}
Zhenzhong Lan, Mingda Chen, Sebastian Goodman, Kevin Gimpel, Piyush Sharma, and
  Radu Soricut. 2019.
\newblock Albert: A lite bert for self-supervised learning of language
  representations.
\newblock \emph{arXiv preprint arXiv:1909.11942}.

\bibitem[{Lewis et~al.(2019)Lewis, Liu, Goyal, Ghazvininejad, Mohamed, Levy,
  Stoyanov, and Zettlemoyer}]{lewis2019bart}
Mike Lewis, Yinhan Liu, Naman Goyal, Marjan Ghazvininejad, Abdelrahman Mohamed,
  Omer Levy, Ves Stoyanov, and Luke Zettlemoyer. 2019.
\newblock Bart: Denoising sequence-to-sequence pre-training for natural
  language generation, translation, and comprehension.
\newblock \emph{arXiv preprint arXiv:1910.13461}.

\bibitem[{Liu et~al.(2019)Liu, Ott, Goyal, Du, Joshi, Chen, Levy, Lewis,
  Zettlemoyer, and Stoyanov}]{liu2019roberta}
Yinhan Liu, Myle Ott, Naman Goyal, Jingfei Du, Mandar Joshi, Danqi Chen, Omer
  Levy, Mike Lewis, Luke Zettlemoyer, and Veselin Stoyanov. 2019.
\newblock Roberta: A robustly optimized bert pretraining approach.
\newblock \emph{arXiv preprint arXiv:1907.11692}.

\bibitem[{Maas et~al.(2011)Maas, Daly, Pham, Huang, Ng, and
  Potts}]{maas2011learning}
Andrew~L Maas, Raymond~E Daly, Peter~T Pham, Dan Huang, Andrew~Y Ng, and
  Christopher Potts. 2011.
\newblock Learning word vectors for sentiment analysis.
\newblock In \emph{Proceedings of the 49th annual meeting of the association
  for computational linguistics: Human language technologies-volume 1}, pages
  142--150. Association for Computational Linguistics.

\bibitem[{Pal and Tople(2020)}]{pal2020transfer}
Bijeeta Pal and Shruti Tople. 2020.
\newblock To transfer or not to transfer: Misclassification attacks against
  transfer learned text classifiers.
\newblock \emph{arXiv preprint arXiv:2001.02438}.

\bibitem[{Raffel et~al.(2019)Raffel, Shazeer, Roberts, Lee, Narang, Matena,
  Zhou, Li, and Liu}]{raffel2019exploring}
Colin Raffel, Noam Shazeer, Adam Roberts, Katherine Lee, Sharan Narang, Michael
  Matena, Yanqi Zhou, Wei Li, and Peter~J Liu. 2019.
\newblock Exploring the limits of transfer learning with a unified text-to-text
  transformer.
\newblock \emph{arXiv preprint arXiv:1910.10683}.

\bibitem[{Rajpurkar et~al.(2016)Rajpurkar, Zhang, Lopyrev, and
  Liang}]{squad2.0}
Pranav Rajpurkar, Jian Zhang, Konstantin Lopyrev, and Percy Liang. 2016.
\newblock \href {https://doi.org/10.18653/v1/D16-1264} {{SQ}u{AD}: 100,000+
  questions for machine comprehension of text}.
\newblock In \emph{Proceedings of the 2016 Conference on Empirical Methods in
  Natural Language Processing}, pages 2383--2392, Austin, Texas. Association
  for Computational Linguistics.

\bibitem[{Ravichander et~al.(2021)Ravichander, Dalmia, Ryskina, Metze, Hovy,
  and Black}]{ravichander2021noiseqa}
Abhilasha Ravichander, Siddharth Dalmia, Maria Ryskina, Florian Metze, Eduard
  Hovy, and Alan~W Black. 2021.
\newblock Noiseqa: Challenge set evaluation for user-centric question
  answering.
\newblock \emph{arXiv preprint arXiv:2102.08345}.

\bibitem[{Ribeiro et~al.(2018)Ribeiro, Singh, and
  Guestrin}]{ribeiro2018semantically}
Marco~Tulio Ribeiro, Sameer Singh, and Carlos Guestrin. 2018.
\newblock Semantically equivalent adversarial rules for debugging nlp models.
\newblock In \emph{Proceedings of the 56th Annual Meeting of the Association
  for Computational Linguistics (Volume 1: Long Papers)}, pages 856--865.

\bibitem[{ROUGE(2004)}]{rouge2004package}
Lin~CY ROUGE. 2004.
\newblock A package for automatic evaluation of summaries.
\newblock In \emph{Proceedings of Workshop on Text Summarization of ACL,
  Spain}.

\bibitem[{Sang and De~Meulder(2003)}]{sang2003introduction}
Erik~F Sang and Fien De~Meulder. 2003.
\newblock Introduction to the conll-2003 shared task: Language-independent
  named entity recognition.
\newblock \emph{arXiv preprint cs/0306050}.

\bibitem[{Seo et~al.(2016)Seo, Kembhavi, Farhadi, and
  Hajishirzi}]{seo2016bidirectional}
Minjoon Seo, Aniruddha Kembhavi, Ali Farhadi, and Hannaneh Hajishirzi. 2016.
\newblock Bidirectional attention flow for machine comprehension.
\newblock \emph{arXiv preprint arXiv:1611.01603}.

\bibitem[{Socher et~al.(2013)Socher, Perelygin, Wu, Chuang, Manning, Ng, and
  Potts}]{socher2013recursive}
Richard Socher, Alex Perelygin, Jean Wu, Jason Chuang, Christopher~D Manning,
  Andrew~Y Ng, and Christopher Potts. 2013.
\newblock Recursive deep models for semantic compositionality over a sentiment
  treebank.
\newblock In \emph{Proceedings of the 2013 conference on empirical methods in
  natural language processing}, pages 1631--1642.

\bibitem[{Subramaniam et~al.(2009)Subramaniam, Roy, Faruquie, and
  Negi}]{Subramaniam2009}
L.~Venkata Subramaniam, Shourya Roy, Tanveer~A. Faruquie, and Sumit Negi. 2009.
\newblock \href {https://doi.org/10.1145/1568296.1568315} {A survey of types of
  text noise and techniques to handle noisy text}.
\newblock In \emph{Proceedings of The Third Workshop on Analytics for Noisy
  Unstructured Text Data}, AND ’09, page 115–122, New York, NY, USA.
  Association for Computing Machinery.

\bibitem[{Sun et~al.(2020)Sun, Hashimoto, Yin, Asai, Li, Yu, and
  Xiong}]{Sun2020AdvBERTBI}
Lichao Sun, Kazuma Hashimoto, Wenpeng Yin, Akari Asai, Jiugang Li, Philip~S.
  Yu, and Caiming Xiong. 2020.
\newblock Adv-bert: Bert is not robust on misspellings! generating nature
  adversarial samples on bert.
\newblock \emph{ArXiv}, abs/2003.04985.

\bibitem[{Taghva et~al.(2000)Taghva, Nartker, Borsack, Lumos, Condit, and
  Young}]{taghva2000evaluating}
Kazem Taghva, Thomas~A Nartker, Julie Borsack, Steven Lumos, Allen Condit, and
  Ron Young. 2000.
\newblock Evaluating text categorization in the presence of ocr errors.
\newblock In \emph{Document Recognition and Retrieval VIII}, volume 4307, pages
  68--74. International Society for Optics and Photonics.

\bibitem[{Wang and Jiang(2016)}]{wang2016machine}
Shuohang Wang and Jing Jiang. 2016.
\newblock Machine comprehension using match-lstm and answer pointer.
\newblock \emph{arXiv preprint arXiv:1608.07905}.

\bibitem[{Wu et~al.(2016)Wu, Schuster, Chen, Le, Norouzi, Macherey, Krikun,
  Cao, Gao, Macherey et~al.}]{wu2016google}
Yonghui Wu, Mike Schuster, Zhifeng Chen, Quoc~V Le, Mohammad Norouzi, Wolfgang
  Macherey, Maxim Krikun, Yuan Cao, Qin Gao, Klaus Macherey, et~al. 2016.
\newblock Google's neural machine translation system: Bridging the gap between
  human and machine translation.
\newblock \emph{arXiv preprint arXiv:1609.08144}.

\bibitem[{Yang et~al.(2019)Yang, Dai, Yang, Carbonell, Salakhutdinov, and
  Le}]{yang2019xlnet}
Zhilin Yang, Zihang Dai, Yiming Yang, Jaime Carbonell, Russ~R Salakhutdinov,
  and Quoc~V Le. 2019.
\newblock Xlnet: Generalized autoregressive pretraining for language
  understanding.
\newblock \emph{Advances in neural information processing systems}, 32.

\end{thebibliography}

\end{document}